\documentclass{article}
\usepackage[preprint]{spconf}
\usepackage{amsmath,graphicx}
\usepackage{hyperref}
\usepackage{cite}
\usepackage{multirow}
\usepackage{amssymb}
\usepackage{booktabs}
\copyrightnotice{\copyright\ IEEE 2021}
\toappear{To appear in {\it Proc.\ ASRU2021,
                    December 13-17, 2021, Cartagena, Colombia}}

\title{Adapting GPT, GPT-2 and BERT Language Models for Speech Recognition}
\name{Xianrui Zheng, Chao Zhang, Philip C. Woodland}
\address{
  Cambridge University Engineering Dept., Trumpington St., Cambridge, CB2 1PZ U.K.\\ 
  \small\texttt{\{xz396, cz277, pcw\}@eng.cam.ac.uk}
}
\begin{document}
\ninept
\maketitle
\begin{abstract}
Language models (LMs) pre-trained on massive amounts of 
text, in particular bidirectional encoder representations from Transformers (BERT), generative pre-training (GPT), and GPT-2, have 
become a key technology for many natural language processing tasks.
In this paper, we present results using fine-tuned GPT, GPT-2, and their combination for automatic speech recognition (ASR). 
Unlike unidirectional LM GPT and GPT-2, BERT is bidirectional whose direct product of the output probabilities is no longer a valid language prior probability. A conversion method is proposed to compute the correct language prior probability based on bidirectional LM outputs in a mathematically exact way. 
Experimental results on the widely used AMI and Switchboard ASR tasks showed that the combination of the fine-tuned GPT and GPT-2 outperformed the combination of three neural LMs with different architectures trained from scratch on the in-domain text by up to a 12\%
relative word error rate reduction (WERR).
Furthermore, on the AMI corpus, the proposed conversion for language prior probabilities enables BERT to 
obtain an extra 3\% relative WERR, and the combination of BERT, GPT and GPT-2 results in further improvements.
\end{abstract}
\section{Introduction}
\label{sec:intro}
Language models (LMs) incorporate linguistic knowledge as the prior probabilities of word sequences, and are crucial for state-of-the-art automatic speech recognition (ASR) systems.
LMs provide a way of leveraging additional text data in ASR \cite{jelinekStatistical1997}. 
Traditional $n$-gram LMs often suffer from data sparsity and are therefore restricted to use only a small number of previous words ($n\leqslant 5$) (i.e. context) when estimating the prior probability of the next word
in a sentences. 
A solution is to build LMs with neural network (NN) models that can more reliably estimate sentence prior probabilities using longer contexts given a certain amount of text training data. 
Alternatively, additional out-of-domain data can be leveraged to improve LM training with limited in-domain data via LM adaptation and transfer learning 
\cite{iyerUsing1997,gangireddyUnsupervised2016,maApproaches2017,salazarMasked2020,shinEffective2019,chiuInnovative2021,liEmpirical2020}.

The feed-forward NN (FNN) was the first NN structure widely studied for language modelling, and can be seen as an NN-based $n$-gram LM \cite{bengioNeural2003,schwenkContinuous2007,parkImproved2010,leStructured2013}.
Later, recurrent neural network (RNN) models and the long short-term memory (LSTM) variant, which can make predictions based on the full history,
were applied to language modelling for ASR \cite{mikolovRecurrent2010,mikolovExtensions2011, sundermeyerLSTM2012,merityRegularizing2017,sundermeyerComparison2013}.

Using an attention-mechanism is an alternative to RNNs for sequence processing \cite{gravesNeural2014,bahdanauNeural2015}.
Transformers, 
a widely used attention-based sequence encoder-decoder model structure, were first proposed for machine translation \cite{vaswaniAttention2017}. The Transformer decoder can be used to build unidirectional LMs for ASR (referred to as Transformer LMs in this paper) \cite{irieLanguage2019}. 
The generative pre-training (GPT) model used the Transformer decoder structure to build an unidirectional LM. The parameters of GPT were first pre-trained 
on very large general text corpora and released to the public \cite{radfordImproving2018}. 
When applied to a specific downstream natural language processing (NLP) task, GPT is often fine-tuned on a small amount of in-domain data. 
This process allows the transfer of linguistic knowledge learned in pre-training to a task with a small amount of task-specific data.
In contrast to GPT, the bidirectional encoder representation from Transformers (BERT) model uses the Transformer encoder structure to build a pre-trained bidirectional LM, 
which leverages both forward and backward context rather than only previous words when computing probabilities \cite{devlinBERT2019}. The success of these models has led to
the study of many other types of pre-trained LM \cite{howardUniversal2018,liuRoBERTa2019,daiTransformerXL2019,petersDeep2018,yangXLNet2020,radfordLanguage2019,lanALBERT2020,raffelExploring2020,brownLanguage2020}.

Despite the wide-spread application of GPT and BERT in NLP and machine learning, there are only a very limited number of studies on their use in ASR \cite{salazarMasked2020,shinEffective2019,chiuInnovative2021,liEmpirical2020}. In this paper, we present ASR results obtained using GPT and GPT-2 that are fine-tuned on in-domain data. The WERs obtained by combining the fine-tuned GPT and GPT-2 LMs outperformed the combination of an FNN LM, an LSTM LM, and a Transformer LM trained only on in-domain data.  
Meanwhile, unlike the unidirectional LMs, simply multiplying the BERT output probabilities over all words in a sentence does not result in a valid sentence prior probability.
A novel method is proposed in this paper that can convert the output probabilities of a bidirectional LM into exact sentence prior probabilities. This method is applied to BERT in our experiments, and is compared to a baseline method developed for the same purpose \cite{chenInvestigating2017,shinEffective2019}.

This paper is organised as follows. Section~\ref{sec:2} reviews Transformer, GPT, GPT-2 and BERT. Section~\ref{sec:3} presents our methods for LM combination and bidirectional LM output probability conversion. The experimental setup for the language model training and speech recognition experiments on the widely used AMI and Switchboard corpora are given in Sec.~\ref{sec:4} and the experimental results are in Sec.~\ref{sec:5}. Finally, conclusions are presented in Sec.~\ref{sec:6}. 

\section{Transformer-based LMs}
\label{sec:2}
This section reviews the Transformer model structure, which is used by GPT, GPT-2, BERT, and the Transformer LM. 

\subsection{Multi-head attention for Transformer}

\begin{figure}[!ht]
\centering
\includegraphics[width=\linewidth]{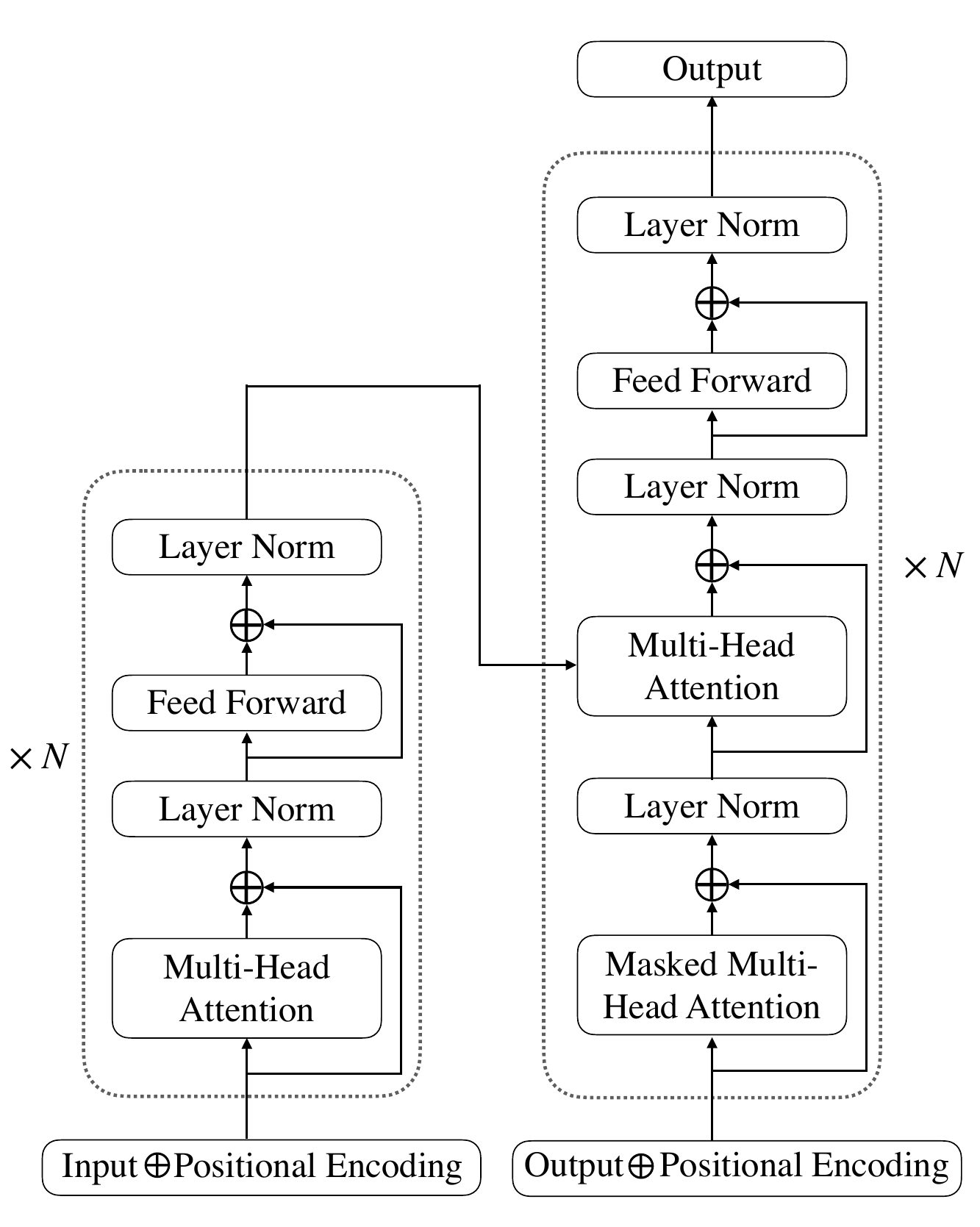}
\caption{Transformer model structure with $N$ encoder blocks (on the left) and $N$ decoder blocks (on the right).}
\label{fig:transformer}
\end{figure}

The Transformer model structure proposed in \cite{vaswaniAttention2017} is shown in Fig.~\ref{fig:transformer}, 
\begin{align*}
    \text{MultiHead}(\mathbf{Q},\mathbf{K},\mathbf{V})&=\text{Concat}(\text{head}_1,\dots,\text{head}_h)\mathbf{W}^\text{O}\\
    \text{head}_i&=\text{Attention}(\mathbf{QW}_i^\text{Q}, \mathbf{KW}_i^\text{K}, \mathbf{VW}_i^\text{V}),\\
    \text{Attention}(\mathbf{Q}, \mathbf{K}, \mathbf{V}) &=\text{softmax}({\mathbf{Q}\mathbf{K}^\text{T}}/{\sqrt{d_k}})\mathbf{V}
\end{align*}
where $\mathbf{K}$, $\mathbf{Q}$, 
and $\mathbf{V}$ refer to the queries, keys, 
and values of the attention mechanism; 
$\text{MultiHead}(\cdot)$ and $\text{Concat}(\cdot)$ refers to multi-head attention and concatenation respectively;
$h$ is the number of heads, 
$\mathbf{W}^\text{O}_i\in\mathbb{R}^{d_\text{model}\times d_\text{model}}$ is a weight matrix of the $i$\,th head, 
and $\mathbf{W}_i^\text{Q}$, $\mathbf{W}_i^\text{K}$ and $\mathbf{W}_i^\text{V}$ have the same dimensions $d_\text{model}\times d_k$, 
where $d_\text{model}$ is the size of input embeddings and $d_k=d_\text{model}/h$. $\text{Attention}(\cdot)$ is termed scaled dot-product attention since it weights the values based on the dot-product of keys and queries.

The difference between Multi-Head Attention and Masked Multi-Head Attention is that the former allows the model to see the future context while the later does not, which are therefore used in the encoder and decoder structures respectively.  
The Feed Forward component consists of two fully-connected (FC) layers with a ReLU function in between. 
The Output component converts the output from the final Transformer decoder block into probability distributions using an FC layer with a softmax function. 
A positional encoding is added to each input embedding to include the order information of the input sequence. 

\subsection{GPT}
\label{sec:2.2}
GPT uses the Transformer decoder structure (shown in the right part of Fig.~\ref{fig:transformer}) \cite{radfordImproving2018}. 
Since the decoder structure is used alone without an encoder, the Multi-Head Attention and Layer Norm (layer normalisation \cite{baLayer2016}) components that are connected to the encoder are removed (in the middle of the right part of Fig.~\ref{fig:transformer}). 
The pre-trained GPT model has 12 Transformer blocks with $d_\text{model}=$768 and 110M parameters. 
The positional encodings are learnt jointly during pre-training.

Regarding GPT, its input is a token sequence $w_{\{i:i+n\}}$ and its output is the probability distributions for the next tokens $w_{\{i+1:i+n+1\}}$. 
Tokenisation uses the byte pair encoding (BPE) with 40,000 subword units \cite{sennrichNeural2016}. 
The maximum token sequence length is 512. 
GPT is pre-trained on the BooksCorpus dataset \cite{zhuAligning2015} for 100 epochs, which consists of one billion words from unpublished books covering many topics. 

\subsection{GPT-2}
\label{sec:2.3}
GPT-2 is the successor to GPT, which  also uses the Transformer decoder structure \cite{radfordLanguage2019}. 
In contrast to GPT, GPT-2 uses 50,257 BPE tokens and places the Layer Norm before the Masked Multi-Head component. An additional Layer Norm is added after the final block. 
The maximum sequence length is increased from 512 to 1024. 
The mini-batch size during pre-training is increased from 64 to 512. 
Four pre-trained GPT-2 models with different numbers of decoder blocks are available. 
The largest one has 48 blocks with $d_\text{model} = 1600$, resulting in a total number of 1.5 billion model parameters. 

The training dataset for GPT-2 is also different to that for GPT. 
By gathering outbound links from Reddit with more than three karma, the resulting training dataset has about ten billion words. 

\subsection{BERT}
Unlike GPT, GPT-2, and Transformer LM, which use the Transformer decoder structure, BERT uses the Transformer encoder structure (see the left part of Fig.~\ref{fig:transformer}) \cite{devlinBERT2019}. Two FC output layers with a Layer Norm component in between are placed 
after the final encoder block. 
The estimation of the output probability of each token relies not only on the previous tokens but also on the future ones, and BERT is therefore a bidirectional LM. 
Two versions of uncased BERT are available. 
The small one has 12 encoder blocks with $d_\text{model}=768$ and the number of parameters is roughly the same as GPT, while the large one has 24 encoder blocks with $d_\text{model}=1024$ and 336M parameters. 

Two tasks are used to pre-train BERT. In the first masked LM task, 15\% of the tokens are replaced by either the symbol \texttt{[MASK]} or a random token. If token $i$ in a sequence is chosen to be replaced, it will have an 80\% probability of being replaced by \texttt{[MASK]}, a 10\% probability of being replaced by a random token and a 10\% probability of remaining unchanged. The goal is to predict the original token. The second task is to predict if a sentence follows another sentence. BERT also uses the BooksCorpus dataset for pre-training. 

\section{Methodology}
\label{sec:3}
\subsection{Combining unidirectional LMs}
For all unidirectional LMs, the training objective is to minimise the log perplexity (PPL) in Eqn. (\ref{equ:logppl}):
\begin{equation}
\begin{aligned}[b]
    \log_2\text{PPL} &= -\frac{1}{T} \log_2 P(w_{1:T}) \\
    &= -\frac{1}{T}\sum\nolimits_{t=1}^{T}\log_2P(w_t|w_{1:t-1}),
    \label{equ:logppl}
\end{aligned}
\end{equation}
where $w_{1:T}$ is a  word sequence with $T$ tokens.
To combine multiple LMs for $n$-best rescoring for ASR, the Covariance Matrix Adaptation Evolution Strategy (CMA-ES) \cite{hansenCompletely2001} can be used.
CMA-ES is used here to optimise a set of LM score scaling factors $\lambda_k$ ($\lambda_k\geqslant0, \forall k$) that minimise the development set WER.
It samples sets of scaling factors from a normal distribution in each iteration and updates the mean and covariance function based on the WER obtained. 
The total score of combining $K$ LMs of a hypothesis is computed by 
\begin{equation}
\begin{aligned}[b]
\text{AMScore} + \sum\nolimits_{k=1}^K \lambda_{k} \log P^{(k)}(w_{1:T}),
\end{aligned}
\label{eqn:2}
\end{equation}
where $P^{(k)}(w_{1:T})$ is the probability estimated by the $k$\,th LM and AMScore is the acoustic model score. 

\subsection{Converting bidirectional LM output probabilities}
\label{sec:exactprob}

In ASR, the sentence prior probability $P(w_{1:T})$ is often calculated using the {chain rule of probability} by
\begin{equation}
P(w_{1:T})=P(w_1)\prod\nolimits_{t=2}^{T}P(w_t|w_{1:t-1}),
\label{eqn:3}
\end{equation}
where $P(w_t|w_{1:t-1})$ requires only the previous tokens to predict the current token and can thus be obtained using a unidirectional LM. Consequently, $P(w_t|w_{1:t-1})$ is often obtained by multiplying the output probabilities of all tokens in the sequence that are produced by  an unidirectional LM. 
For a bidirectional LM where both the previous and future tokens are taken into account, this procedure results in 
\begin{equation}
\Lambda=P(w_1|w_{2:T})P(w_2|w_1,w_{3:T})\ldots P(w_T|w_{1:T-1}).
\label{eqn:4}
\end{equation}
Although $\Lambda\neq P(w_{1:T})$, $\Lambda$ is sometimes directly used to replace $P(w_{1:T})$ in decoding \cite{shinEffective2019,salazarMasked2020,arisoyBidirectional2015}, which we refer to as the modified masked LM (MMLM). 
It was found in \cite{chenInvestigating2017} that the MMLM output distributions are overly-sharp, and hence should be smoothed either by interpolating with the output distributions from other LMs or by using temperature softmax with a temperature factor $\alpha$ ($\alpha<$1): 
\begin{equation}
\text{TempSoftmax}(\mathbf{z})|_{i}=\exp(\alpha z_i)/\sum\nolimits_{j}\exp(\alpha z_j),
\label{eqn:5}
\end{equation}
where $\mathbf{z}$ is the bidirectional LM logit vector. 
It is assumed in \cite{chenInvestigating2017} that $P(w_{1:T})=\Lambda/Z$ where $Z$ is a common normalisation constant calculated over all possible word sequences 
and \cite{heTraining2016} learns similar normalisation terms with noise contrastive estimation. 
In contrast, \cite{irieInvestigation2018} approximates the exact sequence probability using a forward LM, a completion bidirectional LM, and a backward LM. 

Next, the conversion between $\Lambda$ and $P(w_{1:T})$ for a bidirectional LM is discussed.
Based on the definition of conditional probability, $P(w_{1:T})$ can be calculated as
\begin{equation}
\begin{aligned}[b]
P(w_{1:T})&=P(w_1|w_{2:T})P(w_{2:T}) \\
P(w_{1:T})&=P(w_2|w_1,w_{3:T})P(w_1,w_{3:T}) \\ 
&\vdots \\
P(w_{1:T})&=P(w_T|w_{1:T-1})P(w_{1:T-1}).\\
\end{aligned}
\label{eqn:6}
\end{equation}
Multiplying all items in Eqn.~\eqref{eqn:6} together yields
\begin{equation}
P(w_{1:T})=[\Lambda P(w_{2:T})P(w_1,w_{3:T})\ldots P(w_{1:T-1})]^{\frac{1}{T}}.
\label{eqn:7}
\end{equation}
Each term $P(w_{1:t-1},w_{t+1:T})$ is the prior probability obtained by applying Eqn.~\eqref{eqn:7} again over a token string obtained by removing the $t$\,th token from $w_{1:T}$. 
Therefore, Eqn.~\eqref{eqn:7} provides a recursive procedure that converts the bidirectional LM output probabilities into the exact sentence prior probability without using other LMs. This approach to finding the exact sentence prior probability from a bidirectional LM is presented for the first time 
to the best of the authors' knowledge. 
A link between unidirectional and bidirectional LMs can be found by equating the right hand sides of Eqn.~\eqref{eqn:3} and Eqn.~\eqref{eqn:7} (since both equal $P(w_{1:T})$). In this paper, $w_{1:T}$ for the bidirectional model refers to the words in the current sentence and words in other sentences can be given as extra context $C$, \textit{i.e.} all probabilities in this section can be further conditioned on $C$. We omit this extra context $C$ for simplicity. 

\begin{figure}[h!]
\centering
\includegraphics[width=\linewidth]{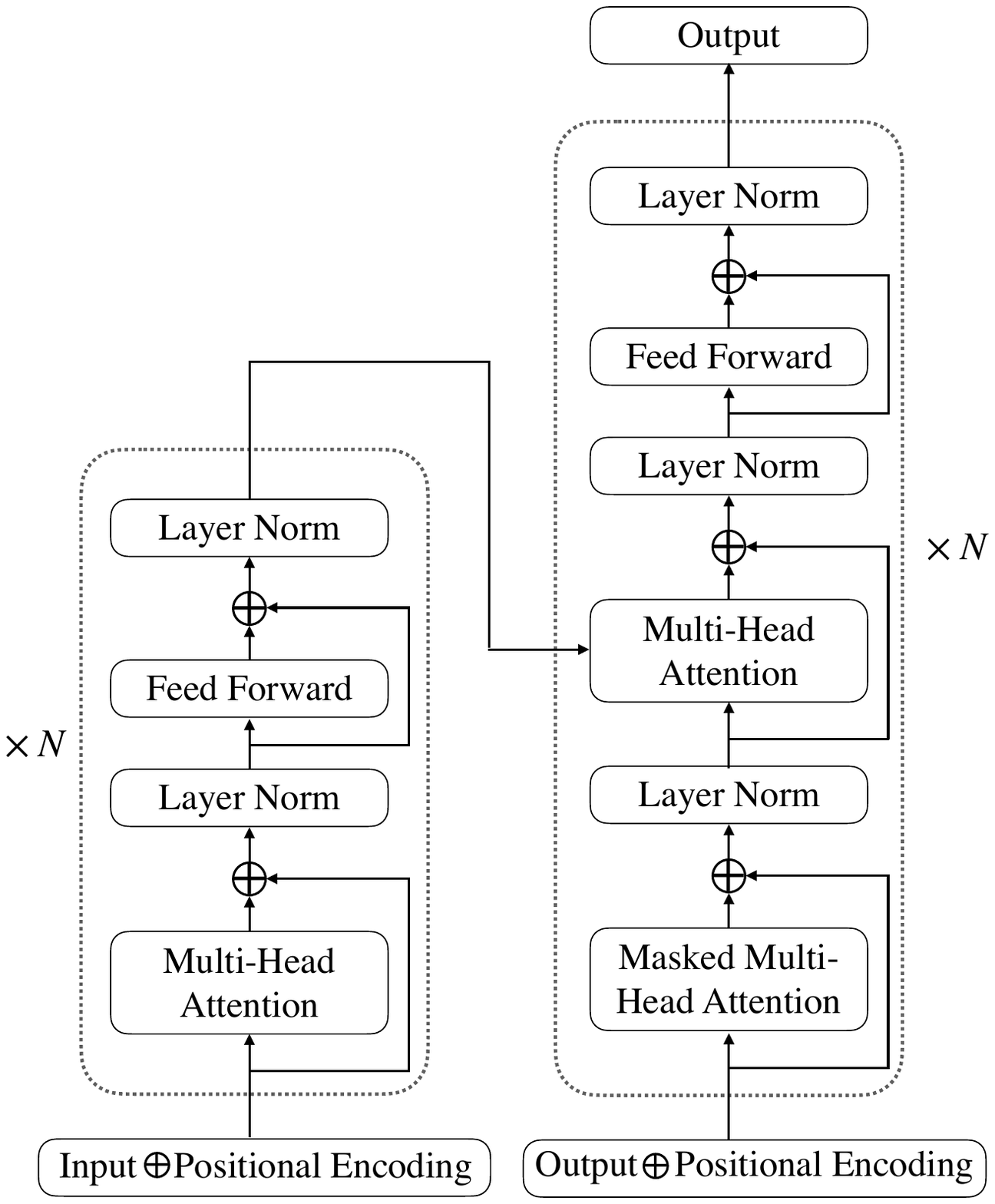}
\caption{An example of the efficient calculation procedure of $P(w_{1:4})$ for a bidirectional LM presented in the form of a finite state acceptor. 
Words to the left within the current sentence are masked in the red path, while words to the right within the current sentence are masked in the blue path. 
}
\label{fig:bieg}
\end{figure}

Rather than applying Eqn.~\eqref{eqn:7} with $T^2$ terms when generating $P(w_{1:T})$, a more efficient calculation procedure is shown in Fig.~\ref{fig:bieg}, which avoids processing repeated token strings by following a specific order of calculation. 
Although this reduces the amount of computation to $0.5T^2+1.5T-1$, it is still computationally impractical when $T$ is large. 
To apply the conversion in practice, we propose an approximation to trade-off between the cost and the context used in the bidirectional LM, which selects $M$ (1$\leqslant M\leqslant T$) items in Eqn.~\eqref{eqn:6} instead of all of them. 
An example with $M=$1 is depicted as the red path in Fig.~\ref{fig:bieg}, which is equivalent to using the bidirectional LM as an unidirectional LM when no tokens from neighbouring sentences are used as extra context. When extra context is provided, the prediction for the target token is conditioned on the tokens on the right and the extra context.

\section{Experimental Setup}
\label{sec:4}
\subsection{Data}
The training set in AMI corpus has 911k word tokens, the dev set (ADev) has 108K and the eval set (AEval) has 102K. 
A 13K word vocabulary was used for NN LMs trained from scratch on AMI. 
Further experiments used the combined training sets from Switchboard and Fisher transcripts (SWB+Fisher), which has a total of 27M words with 30K words in the vocabulary, and are evaluated separately on the SWB and CallHome (CH) parts of the SWB evaluation set \textit{eval2000}. 

\subsection{Acoustic model and 100-best list}

All WERs were obtained using the factorised time-delayed neural network \cite{poveySemiorthogonal2018} acoustic model with residual connections \cite{kreyssigImproved2018}, which was trained using the lattice-free maximum mutual information criterion \cite{poveyPurely2016} following Kaldi recipes \cite{poveyKaldi2011}. 
Neither data augmentation nor speaker adaptation was used for AMI experiments 
while the acoustic model for SWB uses i-vector speaker adaptation and speed perturbation\footnote{The Kaldi pipelines for AMI and SWB are from \cite{manakulAbstractive2020} and \cite{li2021combining} respectively.}. 
A statistical 4-gram model was used to produce the most likely decoding hypotheses for each test utterance represented by a \textit{lattice}. 
The 100-best lists were then extracted for rescoring by the various LMs under investigation.

\subsection{LM Training Procedures}
All LMs trained from scratch used word tokens and optimised by stochastic gradient descent using just the in-domain training data. 
The pre-trained LMs, GPT, GPT-2 and BERT, were fine-tuned using the Adam optimiser \cite{kingmaAdam2015} with only 3 epochs on the in-domain text data. 
All NN LMs were implemented using PyTorch \cite{paszkePyTorch2019} and the pre-trained models were obtained from \cite{wolfTransformers2020}.

\section{Experimental Results}
\label{sec:5}
\begin{table}[t!]
    \centering
    \begin{tabular}{lcccc}
        \toprule
        \textbf{Model} & \textbf{ADev} & \textbf{AEval} & \textbf{SWB} & \textbf{CH} \\
        \midrule
        4-gram  & 19.9 & 20.2 & 8.6 & 17.0 \\
        \midrule
        FNN LM  & 19.4 & 19.5 & 7.9 & 15.8 \\
        LSTM LM  & 18.2 & 17.9 & 6.7 & 13.7 \\
        Transformer LM & 18.4 & 18.4 & 6.6 & 13.7 \\
        \midrule
        F $\oplus$ L $\oplus$ T  & \textbf{17.9} & \textbf{17.7} & \textbf{6.5} & \textbf{13.5} \\
        \bottomrule
    \end{tabular}
    \caption{\%WER on AMI 
    (ADev and AEval) and on \textit{eval2000} (SWB and CH) with different word level LMs trained from scratch to rescore the 100-best lists. 
    F $\oplus$ L $\oplus$ T is the combination of the FNN,  LSTM and  Transformer LMs.}
    \label{tab:baseline}
\end{table}

In our experiments with LMs trained from scratch on only in-domain data, the contexts input to the FNN LM and Transformer LM are 5 words and 72 words on the left respectively, which minimise their perplexities on ADev.
Feeding more context words as input to these two models leads to an increase 
in perplexity on ADev.
A rescored 1-best list was cached during the rescoring process for FNN and Transformer LMs so that they can have enough context by using words from the previous sentences.
For the LSTM LM, the hidden vector of the last word of the rescored best hypothesis is used for rescoring the next sentence $n$-best hypotheses as if giving LSTM LM unlimited context. The same word embedding size of 256 is used for these three LMs. 
We compared our AMI LMs with different depths and sizes, and found for the best-performing models, the FNN, LSTM, and Transformer LMs have 1 FC layer, 1 LSTM layer, and 8 decoder blocks correspondingly. 
For our experiments on SWB+Fisher, we used 2 FC layers, 2 LSTM layers, and 24 decoder blocks for the FNN LM, LSTM LM, and Transformer LM respectively.

Table~\ref{tab:baseline} shows the results rescoring the $100$-best list using a combination of three different in-domain NN LMs. 
The best single in-domain NN LM on AMI (LSTM) is 2.3\% absolute WER lower than that of the 4-gram on AEval, and the combination of the three NN LMs gives an extra 0.2\% absolute reduction on AEval. 
The Transformer LM gives the best WER on SWB and CH, giving 2.0\% and 3.3\% absolute reductions on SWB and CH respectively. 
The combination of three NN LMs further reduces the WER by 0.1\% and 0.2\% absolute over the Transformer LM on SWB and CH respectively. 

\subsection{GPT and GPT-2}

\begin{figure}[t!]
\centering
\includegraphics[width=\linewidth]{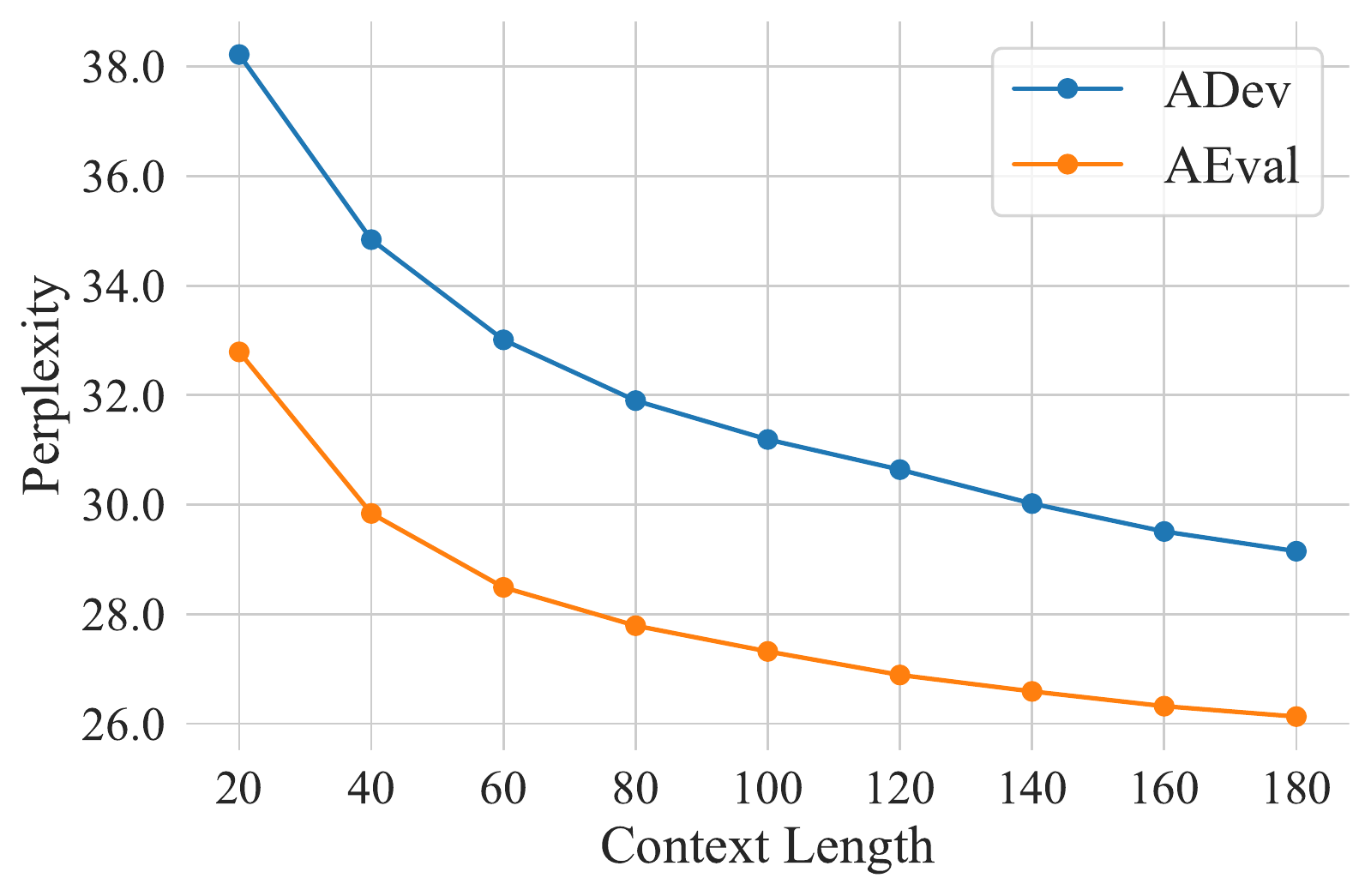}
\vspace{-3mm}
\caption{Preplexity on AMI (ADev and AEval) for GPT fine-tuned with different context lengths. The context length is the number of tokens the model can see for giving an output.}
\label{fig:gptAmiPpl}
\end{figure}

\begin{figure}[t!]
\centering
\includegraphics[width=\linewidth]{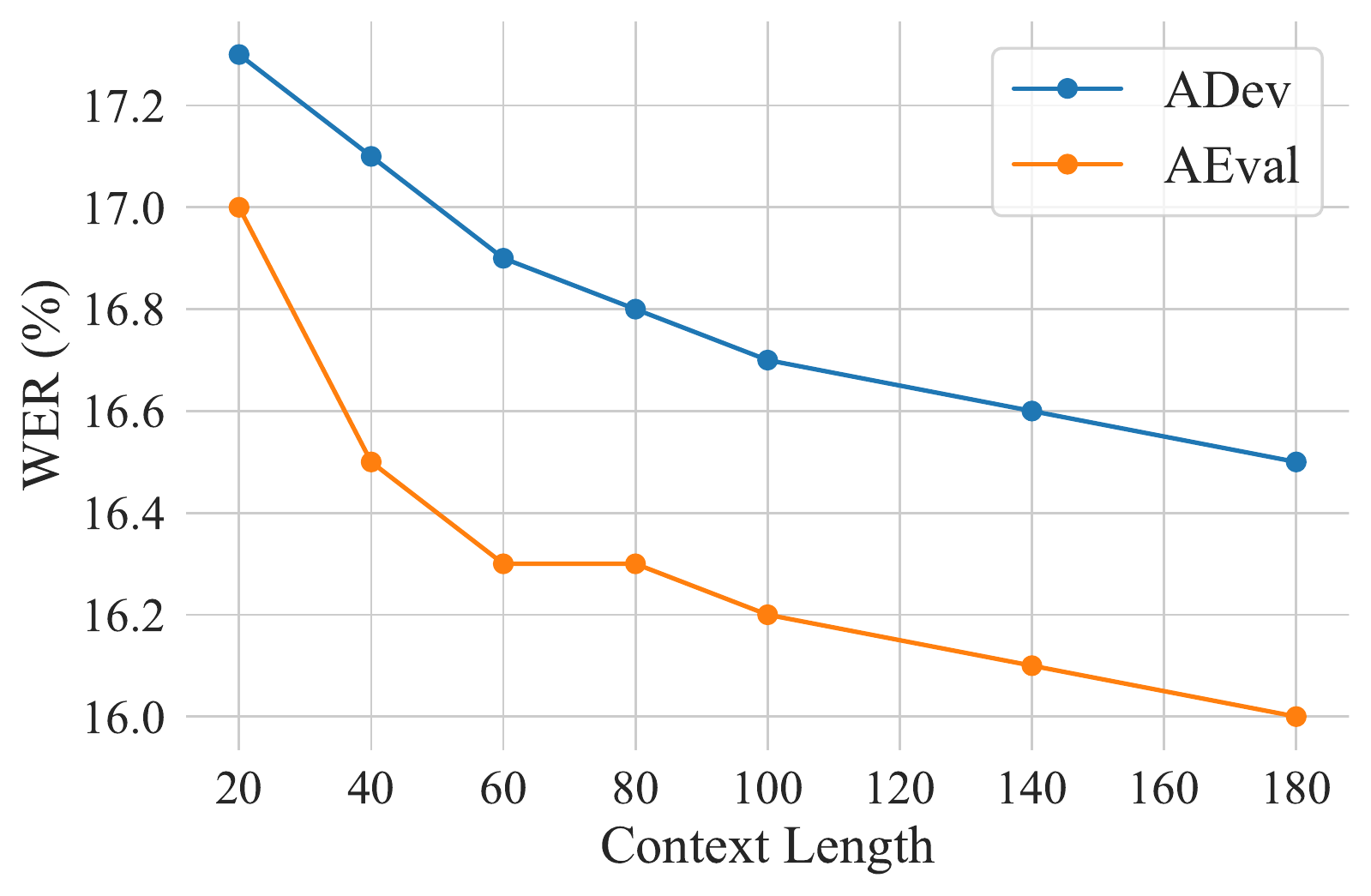}
\vspace{-3mm}
\caption{\%WER on AMI (ADev and AEval) of GPT fine-tuned with different context lengths.}
\label{fig:gptAmiWer}
\end{figure}

The same context length is used for both GPT and GPT-2, and a 1-best list is also cached to serve as context during rescoring.  
To decide the length of context, we fine-tune GPT with different context lengths ranging from 20 to 180 tokens. 
Since GPT was pre-trained on much larger dataset, we believe it can exploit more context than the Transformer LM trained from scratch. 
Fig.~\ref{fig:gptAmiPpl} shows that the perplexity of the fine-tuned GPT drops as the context length increases and the context length of 180 gives the lowest perplexity on both ADev and AEval. 
A similar trend can be found for GPT WERs with different context lengths in Fig.~\ref{fig:gptAmiWer}. The cost is that the time required for rescoring GPT is more than the total time for rescoring the three NN LMs trained from scratch. 

Without fine-tuning, all GPT-2 models shown in Table~\ref{tab:gpt} give lower WERs than F $\oplus$ L $\oplus$ T on AEval. 
The pre-trained 24-block GPT-2 outperforms F $\oplus$ L $\oplus$ T by 1.1\% absolute WER on AEval. 
After fine-tuning, the 24-block GPT-2 gives a 2.0\% absolute WER reduction over F $\oplus$ L $\oplus$ T on AEval. 
Both GPT and the 24-block GPT-2 are adapted to SWB+Fisher and evaluated on \textit{eval2000}, the fine-tuned GPT outperforms F $\oplus$ L $\oplus$ T by 0.2\% and 0.4\% absolute WER on SWB and CH respectively. 
GPT-2 further reduces the WERs by 0.1\% and 0.3\% absolute on SWB and CH respectively. 

Next, the complementarity of GPT and GPT-2 are investigated by linearly interpolating the scores derived from the 24-block GPT-2 and GPT.
The resulting WERs decreased by 0.2\% absolute on ADev and 0.1\% absolute on AEval, compared to the WER using the fine-tuned 24-block GPT-2 only. 
Combining the fine-tuned GPT and 24-block GPT-2 also showed an extra 0.1\% absolute WER reduction on SWB and CH over a single 24-block GPT-2, reaching WERs of 6.1\% and 12.7\%\footnote{These WERs on SWB and CH eval2000 were, at the time that this paper was submitted, lower than the best result in \url{https://github.com/syhw/wer_are_we} (6.8\%/14.4\% on SWB/CH) when a single acoustic model is used and trained on the Switchboard 300 hour set only \cite{wangInvestigation2020}.}, which gives 2.5\% and 4.3\% absolute WER reductions over 4-grams on SWB and CH. 
We found that combining the in-domain only word-level NN LMs with the fine-tuned GPT and GPT-2 could not achieve a better WER. 

\begin{table}[t!]
    \centering
    \begin{tabular}{lccccc}
        \toprule
        \textbf{Model} & \textbf{FT} & \textbf{ADev} & \textbf{AEval} & \textbf{SWB} & \textbf{CH}\\
        \midrule
        F $\oplus$ L $\oplus$ T  & - & 17.9 & 17.7 & 6.5 & 13.5 \\
        \midrule
        \multirow{2}{*}{GPT} & $\times$  & 19.2 & 18.9 & - & - \\
         & $\surd$  & 16.5 & 16.0 & 6.3 & 13.1 \\
        \midrule
        GPT-2  & $\times$ & 17.7 & 17.3 & - & - \\
        (12 blocks) & $\surd$ & 16.4 & 16.0 & - & -  \\
        \midrule
        GPT-2   & $\times$ & 17.1 & 16.6 & - & -  \\
        (24 blocks) & $\surd$ & 16.2 & 15.7 & 6.2 & 12.8 \\
        \midrule
        GPT $\oplus$ GPT-2 & $\boldsymbol\surd$ & \textbf{16.0} & \textbf{15.6} & \textbf{6.1} & \textbf{12.7} \\
        \bottomrule
    \end{tabular}
    \caption{\%WER on AMI (ADev and AEval) and on \textit{eval2000} (SWB and CH) 
    with GPT/GPT-2. ``FT'' indicates if the pre-trained model is fine-tuned on in-domain data. F $\oplus$ L $\oplus$ T is from Table~\ref{tab:baseline}, which is trained on in-domain data only. The last line combines GPT with the 24-block GPT-2.}
    \label{tab:gpt}
\end{table}

\subsection{BERT}

BERT takes much longer to train than uni-directional models since every word needs to be masked once and fed into the model with left and right context. The experiments using BERT start without using context from other sentences, which is the same as the work in \cite{shinEffective2019}. 
Instead of prepending \texttt{[CLS]} and appending \texttt{[SEP]} to each sentence as in \cite{salazarMasked2020}, we only append a period to the end of each sentence since there is no punctuation in the nbest list. 
We found that adding \texttt{[CLS]} and \texttt{[SEP]} to each sentence during fine-tuning did not improve the rescoring result. 
Since we will later allow the model to use the context from neighbouring sentences, having these two special tokens for every sentence reduces the number of tokens from transcriptions under the same context length. 

Similar to the MMLM, our proposed method in Section~\ref{sec:exactprob} also replaces the word we want to predict by the special token \texttt{[MASK]}, while the attention mask is used to control the context. 
Table~\ref{tab:bertNoContext} compares our proposed method with MMLM without letting the model see any tokens outside the current sentence. 
To match the computation needed for MMLM, we use the simplest case of our method by setting $M=1$, where the attention mask allows BERT to see only the future tokens in the current sentence. 
Our method with $M=2$ computes both the red path and the blue path in Fig~\ref{fig:bieg}. 
When there is no fine-tuning, the performance is better using MMLM. 
This is expected since the MMLM method is better aligned to the original \texttt{Masked LM} pre-training task. 
Compared to MMLM, our methods gives lower WERs after fine-tuning, showing a 0.7\% absolute WER reduction on AEval when using $M=1$. We also set $\alpha=0.7$ (as in \cite{chenInvestigating2017}) for decoding BERT using the MMLM method and this gives a 0.8\% absolute WER reduction on AEval after fine-tuning, compared to the MMLM result with $\alpha=1$. 
Our method also benefits from applying smaller $\alpha$. Compared to MMLM with $\alpha=0.7$, setting $M=1$ with $\alpha=0.7$ gives a 0.2\% absolute WER reduction on AEval, and $M=2$ with $\alpha=0.7$ gives a further 0.2\% absolute reduction. 

\begin{table}[t!]
\centering
\begin{tabular}{lcccc}
\toprule
\textbf{Method} & $\boldsymbol{\alpha}$ & \textbf{FT} & \textbf{ADev} & \textbf{AEval} \\
\midrule
\multirow{3}{*}{MMLM} 
& 1 & $\times$ & 23.3 & 23.2 \\
& 1 & $\surd$ & 18.8 & 18.9 \\ 
& 0.7 & $\surd$ & 18.1 & 18.1 \\ 
\midrule
\multirow{3}{*}{Ours ($M=1$)} 
& 1 & $\times$ & 25.8 & 26.0 \\
& 1 & $\surd$ & 18.3 & 18.2 \\ 
& 0.7 & $\surd$ & 18.2 & 17.9 \\ 
\midrule
Ours ($M=2$) & 0.7 & $\surd$ & \textbf{18.0} & \textbf{17.7} \\ 
\bottomrule
\end{tabular}
\caption{\%WER on AMI (ADev and AEval)
with the 12-block uncased BERT using different methods to calculate the sentence probability without further context beyond the current sentence. “FT” indicates if the pre-trained model is fine-tuned on in-domain data. ``Ours'' applies Eqn.~\eqref{eqn:7} with different values of $M$.}
\label{tab:bertNoContext}
\end{table}

\begin{table}[ht!]
\centering
\begin{tabular}{lcccc}
\toprule
\textbf{Method} & \textbf{LC} & \textbf{RC} & \textbf{ADev} & \textbf{AEval} \\
\midrule
\multirow{5}{*}{MMLM} 
& 0 & 0 & 18.1 & 18.1 \\
\cmidrule(l){2-5}
& 50 & 0 & 17.7 & 17.8 \\
& 100 & 0 & 17.6 & 17.6 \\
\cmidrule(l){2-5}
& 50 & 20 & 17.5 & 17.6 \\
& 100 & 100 & 17.5 & 17.5 \\
\midrule
\multirow{8}{*}{Ours ($M=1$)}
& 0 & 0 & 18.2 & 17.9 \\
\cmidrule(l){2-5}
& 50 & 0 & 17.5 & 17.3 \\
& 100 & 0 & 17.5 & 17.3 \\
\cmidrule(l){2-5}
& 50 & 20 & 17.2 & 17.0 \\
& 50 & 50 & 17.3 & 17.1 \\
& 100 & 50 & 17.2 & 17.1 \\
& 100 & 100 & 17.2 & 17.1 \\
\midrule
\multirow{3}{*}{Ours ($M=2$)}
& 0 & 0 & 18.0 & 17.7 \\
\cmidrule(l){2-5}
& 50 & 20 & \textbf{17.0} & \textbf{16.9} \\
& 50 & $20^{\dagger}$ & 16.8 & 16.7 \\
\midrule
\multicolumn{2}{l}{GPT $\oplus$ GPT-2 $\oplus$ BERT} & & \textbf{15.9} & \textbf{15.5} \\
\bottomrule
\end{tabular}
\caption{\%WER on AMI (ADev and AEval)
with the fine-tuned 12-block uncased BERT using different methods to calculate the sentence probability with context. 
``LC'' and ``RC'' indicates context length on the left and right respectively. $\alpha$ is set to 0.7 for rescoing using BERT models. $20^\dagger$ means 20 future tokens from the reference. 
The last line combines GPT, 24-block GPT-2 and BERT model using the highlighted setup of our method with $M=2$.
}
\label{tab:bert}
\end{table}

Unlike GPT, GPT-2 and the NN LMs trained from scratch, which only use context from previous sentences, the context for BERT can come from both previous and future sentences. 
A rescored 1-best list is cached to serve as the left context. Since the future hypotheses cannot be rescored before the current hypothesis, the right context of the current hypothesis uses the 1-best hypotheses from the original 100-best list. 
For our method with $M=1$, Table~\ref{tab:bert} shows that having 50 tokens on the left gives 0.7\% and 0.6\% absolute WER reductions on ADev and AEval respectively. 
Increasing the length of the left context to 100 does not give further WER reductions. 
However, using 50 tokens on the left and 20 on the right outperforms using 100 tokens on the left without context on the right, yielding a 0.3\% absolute WER reduction on both ADev and AEval, which is the best result with $M=1$. 

Compared to the best result using MMLM, which is when both sides have 100 tokens, our method with $M=1$ gives a 0.3\% absolute WER reduction on ADev and 0.5\% reduction on AEval. 
Applying the best setup for our method with $M=1$ on $M=2$ achieves another 0.2\% and 0.1\% absolute WER reduction on ADev and AEval respectively.
The second line for our method with $M=2$ in Table~\ref{tab:bert} shows that if we improve the quality of the right context by using the reference transcriptions instead of the 1-best hypotheses from the original 100-best list, the absolute WERs on ADev and AEval can be reduced by 0.2\%. Future work could investigate approaches to improve the quality of the right context without using the reference. 
Since our method with $M=2$ using the highlighted setup in Table~\ref{tab:bert} gives the best result among all methods for fine-tuning BERT, it is then combined with the GPT and the 24-block GPT-2. 
This combination gives 0.1\% absolute WER reductions on both ADev and AEval compared to the combination of GPT and GPT-2 only, showing that bidirectional LMs are complementary to unidirectional LMs.

\section{Conclusions}
\label{sec:6}
This paper investigates the use of both unidirectional and bidirectional NN LMs with different model architectures and training strategies for ASR with speech recognition experiments on both the widely used AMI  and Switchboard corpora.  
Regarding unidirectional LMs, it was shown that fine-tuning the pre-trained GPT and GPT-2 LMs on small and medium sized in-domain datasets outperformed the combination of FNN LM, LSTM LM, and Transformer LM that are all trained from scratch with only in-domain data. The WERs can be further reduced by combining the fine-tuned GPT and GPT-2. Compared to the 4-gram baseline, combining GPT and GPT-2 gives 4.6\% absolute reductions in WER on the AMI eval set and 2.5\%/4.3\% on the Switchboard/Call Home portions of the eval2000 data. Hence, rather than building new NN LMs with task-specific data for ASR, it can be better and perhaps faster to fine-tune the existing unidirectional LMs that have been pre-trained on huge amounts of out-of-domain data. However, these large pre-trained models require much longer rescoring time, so it is worth considering some optimisation methods during inference. Regarding bidirectional LMs, a conversion method is proposed that can compute the exact sentence prior probabilities required by ASR based on the output proabilities produced by a bidirectional LM, such as BERT. The lowest WERs were achieved by further combining fine-tuned GPT, GPT-2, and BERT together.

\bibliographystyle{IEEEbib}
\bibliography{refs}

\end{document}